\documentclass[conference]{IEEEtran}
\IEEEoverridecommandlockouts
\usepackage{cite}
\usepackage{float}
\usepackage{amsmath,amssymb,amsfonts}
\usepackage{algorithmic}
\usepackage{graphicx}
\usepackage{textcomp}
\usepackage{xcolor}
\usepackage{soul}
\usepackage{hyperref}

\def\BibTeX{{\rm B\kern-.05em{\sc i\kern-.025em b}\kern-.08em
    T\kern-.1667em\lower.7ex\hbox{E}\kern-.125emX}}

\title{Summary Paper:
Use Case on Building Collaborative Safe Autonomous Systems  \\A Robotdog for Guiding Visually Impaired People
}

\author{\IEEEauthorblockN{Aman Malhotra}
\IEEEauthorblockA{\textit{Chair of Embedded Systems} \\
\textit{TU Dortmund University}\\
Dortmund, Germany \\
aman.malhotra@tu-dortmund.de}
\and
\IEEEauthorblockN{Selma Saidi}
\IEEEauthorblockA{\textit{Chair of Embedded Systems} \\
\textit{TU Dortmund University}\\
Dortmund, Germany \\
selma.saidi@tu-dortmund.de}
}

\begin{document}

\maketitle

\begin{abstract}
This is a summary paper of a use case of a Robotdog dedicated to guide visually impaired people in complex environment like a smart intersection. In such scenarios, the Robotdog has to autonomously decide whether it is safe to cross the intersection or not in order to further guide the human. We leverage data sharing and collaboration between the  Robotdog and other autonomous systems operating in the same environment. We propose a system architecture for autonomous systems through a separation of a collaborative decision layer, to enable collective decision making processes, where data about the environment, relevant to the Robotdog decision, together with evidences for trustworthiness about other systems and the environment are shared.




\end{abstract}

\section{What is the Problem we are Solving?}
\begin{figure}[]
	\centering
  \includegraphics[scale = 0.20]{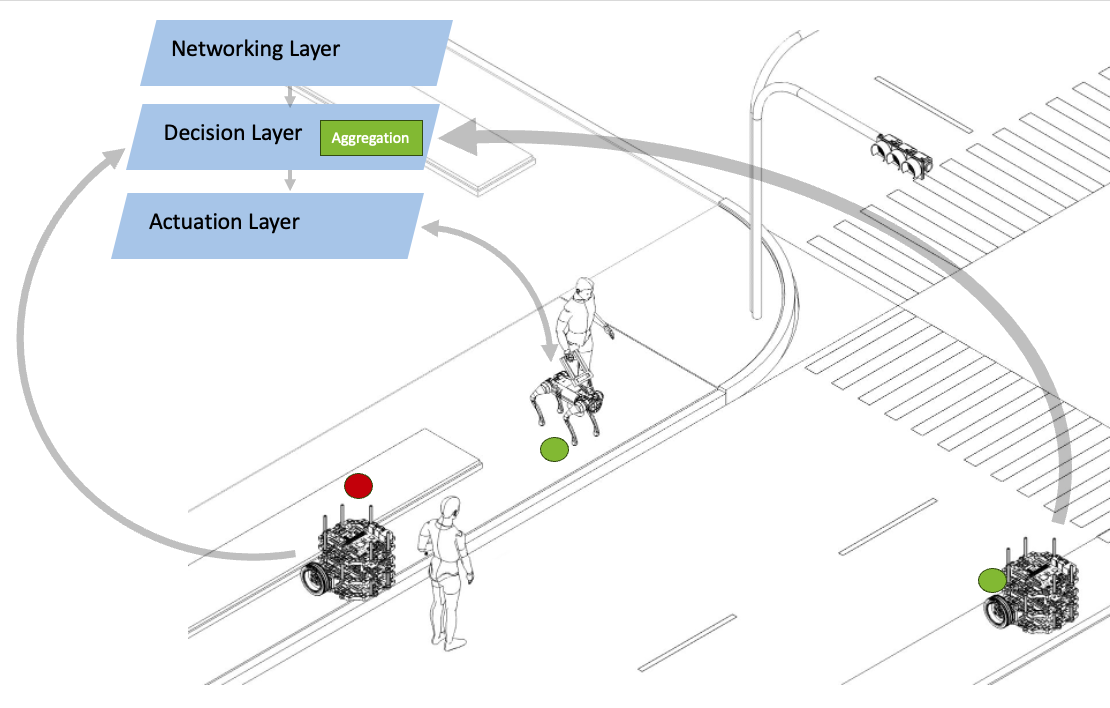}
	\caption{Sketch of the Robotdog operating in a smart intersection. Layers of system architecture for enabling collaborative decision making are depicted. A full demo can be found on the following link: 
\href{https://tinyurl.com/tudrobotdog}{https://tinyurl.com/tudrobotdog}}

	\label{fig:smart-intersection-sketch}
\end{figure}

Autonomous systems are becoming an integral part of many application domains, including mobility and human-assistive robotics. Human-assistive robotics, particularly for social care, offer immense potential in aiding individuals with disabilities in their daily life. 
However, ensuring safe and correct behavior of such systems in dynamic and complex environments remains a significant challenge. 
We consider the example of Robotdogs guiding blind and visually impaired people. 
Similarly to standard guide dogs~\cite{guidedog},
requirements on safety are crucial when considering the design of those systems. These requirements include good traffic awareness and safety, identifying obstacles or safety hazards in their path, and guiding the handler safely around them, including in public environments. We focus on developing decision-making
processes for safety critical autonomous systems, particularly in scenarios such as smart intersection depicted in~\ref{fig:smart-intersection-sketch}), where the Robotdog needs to
autonomously determine whether it is safe to cross the
intersection without colliding with other (autonomous)
systems or pedestrians. %
We propose a collaborative framework enabling autonomous systems to leverage collective decision-making processes and increase reliability. By allowing autonomous systems to share information and aggregate data, we aim to reduce (perception) errors and improve decision outcomes. 
For that, we propose a system architecture supporting collaborative approaches for assuring autonomy and good decision making. Two key aspects from a design perspective are i) separation between the decision layer and the rest of sensing and actuating functionalities, this is critical as the system needs to decide about which information can be shared and will contribute to the collective decision making, ii) how to aggregate such information to improve decision making collaboratively? several aggregation rules and semantics may lead to different degrees of increased reliability. 

\section{What is the Main Idea of the Solution?}
Autonomous systems are often designed with multiple embedded components for perception, trajectory planning, and decision-making. By leveraging collaboration, we build a dynamic and networked distributed system where nodes are autonomous. We consider representative scenarios like the one depicted in Fig,~\ref{fig:smart-intersection-sketch}.  The scenario involves a road intersection with several autonomous systems (e.g., TurtleBots) and a human-assistive Robotdog. Every autonomous system is equipped with different types of sensors (Lidar, optical and RGB camera) and is present at a different location of the intersection, thereby at different distances from a potential obstacle. Systems are therefore purposely designed with different perception and actuation capabilities.  Enabling collaboration with the Robotdog assisting a human in crossing the intersection poses real-time and safety challenges. 
We propose a system architecture separating three layers, i) Sensing and Networking Layer, ii) Decision Layer, and iii) Actuation Layer, which is fundamental for software development and safe decision-making. Each layer can be performed individually or in collaboration.  

\begin{itemize}
    \item Sensing and Networking Layer: The role of this layer is to sense the environment and share knowledge (data points in space) for decision making. Sensing happens in a distributed manner on each node (in our case separately on the Robotdog and TurtleBots).  
    We use WiFi 802.11\footnote{Other wireless communication networks like 5G/6G can also be used.}, to enable mesh networking to communicate data and helps in building and establishing multi-node setup. In our approach we consider one master (the Robotdog) that acts as the decision making node, the rest are slave nodes.  We use ROS~\cite{ROS} designed to communicate at a master slave setup and
    allows a common subscriber - publisher approach to perform  data sharing. 
\item Decision Layer: 
The output of the decision layer is a safe trajectory computed for every autonomous system, and more critically for the Robotdog guiding the human. 
For that, proper detection of objects and obstacles in the environment is crucial. Every autonomous systems is restricted in terms of sensing and perception to the field of view of its sensors and this is what we would like to improve using a \emph{collaborative decision layer}. 
Note that we do not aim here at performing a complete collaborative perception system (e.g., using sensors fusion) or a complete collaborative planning software stack, we rather aim at providing a systemic approach in enabling collaboration by sharing and aggregating information, critical for good decision making. 
Let us consider for every autonomous system the results of perception system after classification (e.g., in a simple binary form for statements like "pedestrian has been detected"). Such claims are shared through the sensing and networking layer and passed on to the decision layer, together with information such as sensors quality of the autonomous system emitting the claim. These quality attributes act as evidence for trustworthiness for the claim. 
Ranking for autonomous systems based on their quality of attributes, as suggested in~\cite{saidi2023collective}, to decide whether an autonomous system is trustworthy or not and therefore their claims can be trusted by the Robotdog is proposed. In the simple example of Fig.~\ref{fig:smart-intersection-sketch}, the Robotdog together with another TurtlBot based on their sensors information conclude that no pedestrian is detected (indicated with a green dot), while another TurtlBot conclude the opposite (indicated with a red dot). Since the second TurtleBot has better quality of sensors and is closer to the pedestrian, it can be considered as the most trustworthy (i.e., expert) in the group, the Robotdog decides to stop and not to traverse the intersection\footnote{A more elaborated example and demo is in the indicated link of the Figure.}. The decision layer requires i)comparing symbolically different types of sensors and their quality to perform a ranking, ii) collect data in a timely manner for comparison, For that we use the notion of sessions or frames during which sensor data observation is facilitated. 

\item Actuation Layer: this layer considers the actuation context and output of the decision layer to activate the Robotdog. Making sure autonomous systems acte precisely and reliably. This requires a sequential process of steps to solve the data to trigger the actuation. This actuation trigger is required to be initiated with a session start and end.  Actuation sequencing is prioritized based on device IDs, with safety criticality managed through sequential sessions, enabling timely updates to actuation processes.
\end{itemize}

\section{What is the Expected Impact?}
\paragraph{Scientific impact}
Traditional designs of autonomous systems focus on the development of Sense-Decide-Act blocks in classical cyber-physical systems for functionality like perception and maneuvering, using increasing support from learning enabled components. This might be sufficient for the Robotdog to operate autonomously but with no (or little) safety guarantees in complex environments. We believe, as previously advocated in~\cite{our-ieee} through the supervision layer, that there is a need for more systems that control safe decision making at operation time. 
Defining appropriate aggregation and collaboration rules and deciding which data is relevant to aggregate becomes key in increasing trustworthiness  and safety. 



\paragraph{Social Impact} 
There is a promise that collaboration, enabled by infrastructure like in smart intersections, 
help builds trustworthy models of the environment of operation~\cite{liu2022ieee}. The proposed work in this paper goes in line with this vision that can be further used not just for the development of automated driving but also for human-assistive robotics in outdoor scenarios. This will allow more autonomy for visually impaired people and more safety since standard guide dogs are very costly and trained for a specific sequence of places. Changes in the environment will therefore require guide dogs to be trained further. We believe that autonomous systems like the proposed Robotdog can offer more efficient and less costly alternative solutions for visually impaired people. 

\bibliographystyle{unsrt}
\bibliography{biblio}

\end{document}